\documentclass[11pt]{article}

\usepackage{times,epsfig,pictex,epic}
\usepackage{url}
\usepackage{examples}
\usepackage{avm}
\usepackage{ecltree}
\usepackage[T1]{tipa}
\usepackage{fancybox}\cornersize{1}
\def\colname#1{{\bf #1}}
\def\term#1{{\bfseries #1}}
\def\ling#1{{\itshape #1}}
\usepackage{natbib}
\bibpunct[:]{(}{)}{;}{a}{,}{,}

\def\bez#1#2#3{{\setlength{\unitlength}{.75mm}%
\begin{picture}(3,6)\bezier{#1}(0,#2)(2,#2)(3,#3)\end{picture}}}
\def\qbez#1#2{{\setlength{\unitlength}{.7mm}%
\begin{picture}(3,6)\qbezier(0,#1)(2,#1)(3,#2)\end{picture}}}

\def\d#1#2{\bez{4}{#1}{#2}}  
\def\l#1#2{\qbez{#1}{#2}}    

\title{Phonology}
\author{Steven Bird\\
University of Pennsylvania}
\date{2002}

\begin{document}
\maketitle

\abstract{
Phonology is the systematic study of the sounds used in language,
their internal structure, and their composition into syllables,
words and phrases.
Computational phonology is the application of formal and computational
techniques to the representation and processing of phonological
information.  This chapter will present
the fundamentals of descriptive phonology along with a brief overview
of computational phonology.
}

\section{Phonological contrast, the phoneme, and distinctive features}

There is no limit to the number of distinct sounds that can be
produced by the human vocal apparatus.  However, this infinite
variety is harnessed by human languages into \term{sound systems}
consisting of a few dozen language-specific categories, or
\term{phonemes}.  An example of an English phoneme is \ling{t}.
English has a variety of \ling{t}-like sounds,
such as the aspirated \ling{t\textsuperscript{h}} of \ling{ten}
the unreleased \ling{t\textcorner} of \ling{net}, and
the flapped \ling{\textfishhookr} of \ling{water} (in some dialects).
In English, these distinctions are not used to differentiate
words, and so we do not find pairs of
English words which are identical but for their use of
\ling{t\textsuperscript{h}} versus \ling{t\textcorner}.
(By comparison, in some other languages, such as
Icelandic and Bengali, aspiration is contrastive.)
Nevertheless, since these sounds (or \term{phones}, or \term{segments})
are phonetically similar,
and since they occur in \term{complementary distribution}
(i.e. disjoint contexts) and cannot differentiate words in
English, they are all said
to be \term{allophones} of the English phoneme \ling{t}.

Of course, setting up a few allophonic variants for each
of a finite set of phonemes does not account for the infinite
variety of sounds mentioned above.  If one were to record
multiple instances of the same utterance by the single speaker,
many small variations could be observed in loudness, pitch, rate,
vowel quality, and so on.  These variations arise because speech is a
motor activity involving coordination of many independent
articulators, and perfect repetition of any utterance
is simply impossible.  Similar variations occur between different
speakers, since one person's vocal apparatus is different to the
next person's (and this is how we can distinguish people's voices).
So 10 people saying \ling{ten} 10 times each will produce 100
distinct acoustic records for the \ling{t} sound.  This diversity
of tokens associated with a single type is sometimes referred to as
\term{free variation}.

\begin{figure}[t]
\centerline{\epsfig{figure=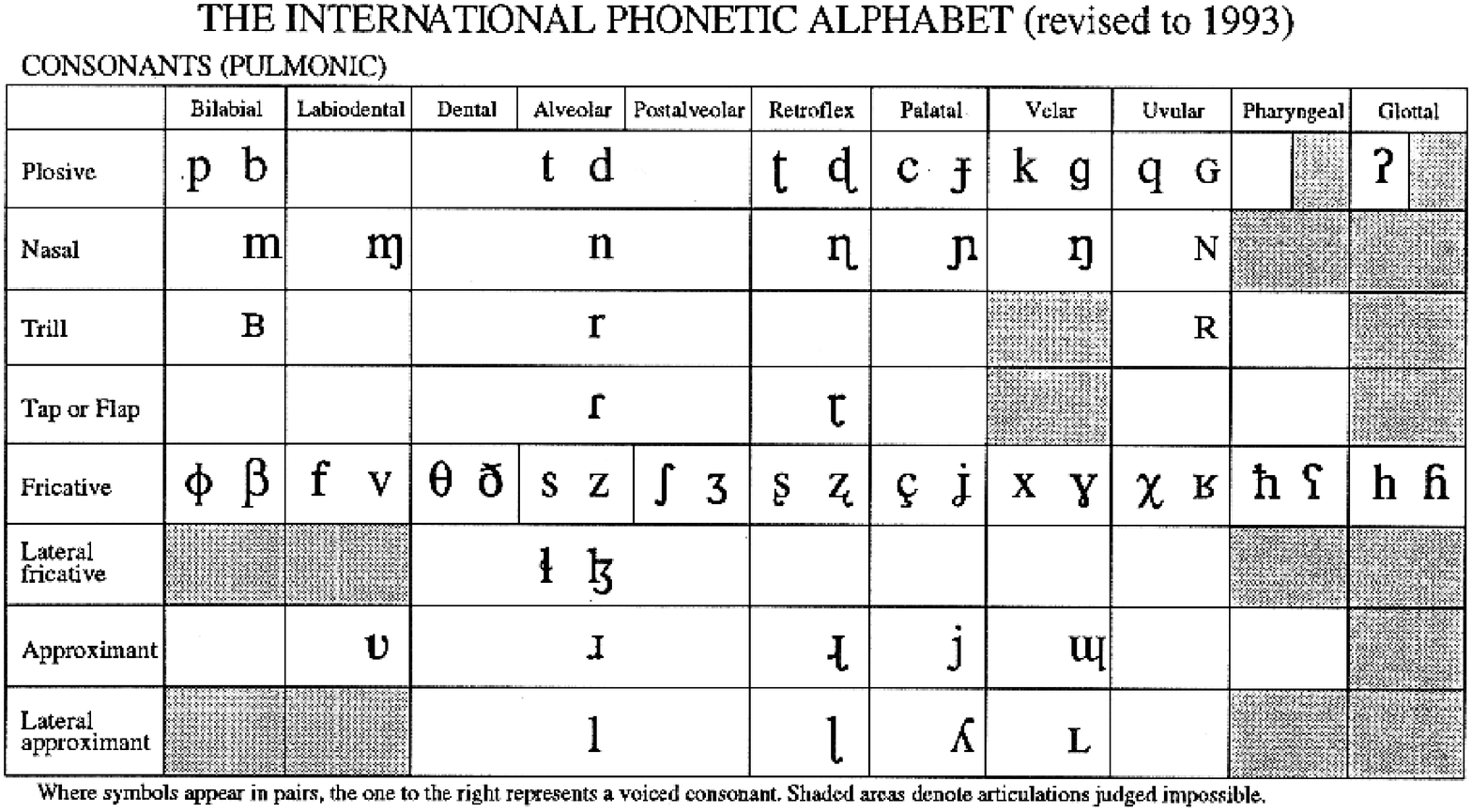,width=\linewidth}}
\caption{Pulmonic Consonants from the International Phonetic Alphabet}
\label{fig:ipa}
\end{figure}

Above, the notion of phonetic similarity was used.  The primary way
to judge the similarity of phones is in terms of their
\term{place} and \term{manner} of articulation.  The consonant chart
of the International Phonetic Alphabet (IPA) tabulates phones in
this way, as shown in Figure~\ref{fig:ipa}.  The IPA provides symbols for all
sounds that are contrastive in at least one language.

The major axes of this chart are for place of articulation (horizontal),
which is the location in the oral cavity of the primary constriction,
and manner of articulation (vertical), the nature and degree of that
constriction.  Many cells of the chart contain two consonants, one
\term{voiced} and the other \term{unvoiced}.  These complementary
properties are usually expressed as opposite values of a
\term{binary feature} [$\pm$voiced].

A more elaborate model of the similarity of phones is provided by
the theory of \term{distinctive features}.  Two phones are considered
more similar to the extent that they agree on the value of their
features.  A set of distinctive features and their values for
five different phones is shown in (\ref{ex:distinctive}).
(Note that many of the features have an extended technical definition,
for which it is necessary to consult a textbook.)

\begin{examples}
\item\label{ex:distinctive}
\begin{tabular}[t]{lccccc}
              & t & z & m & l & i \\
anterior      & $+$ & $+$ & $+$ & $+$ & $-$ \\
coronal       & $+$ & $+$ & $-$ & $+$ & $-$ \\
labial        & $-$ & $-$ & $+$ & $-$ & $-$ \\
distributed   & $-$ & $-$ & $-$ & $-$ & $-$ \\
consonantal   & $+$ & $+$ & $+$ & $+$ & $-$ \\
sonorant      & $-$ & $-$ & $+$ & $+$ & $+$ \\
voiced        & $-$ & $+$ & $+$ & $+$ & $+$ \\
approximant   & $-$ & $-$ & $-$ & $+$ & $+$ \\
continuant    & $-$ & $+$ & $-$ & $+$ & $+$ \\
lateral       & $-$ & $-$ & $-$ & $+$ & $-$ \\
nasal         & $-$ & $-$ & $+$ & $-$ & $-$ \\
strident      & $-$ & $+$ & $-$ & $-$ & $-$
\end{tabular}
\end{examples}

Statements about the distribution of phonological information,
usually expressed with rules or constraints, often apply
to particular subsets of phones.  Instead of listing these
sets, it is virtually always simpler to list two or three
feature values which pick out the required set.
For example [+labial,--continuant] picks out
\ling{b}, \ling{p}, and \ling{m}, shown in the top left
corner of Figure~\ref{fig:ipa}.
Sets of phones which can be picked out in this way
are called \term{natural classes}, and phonological
analyses can be evaluated in terms of their reliance
on natural classes.  How can we express these analyses?
The rest of this chapter discusses some key approaches to
this question.

Unfortunately, as with any introductory chapter like this one,
it will not be possible to cover many important topics of
interests to phonologists, such as
acquisition, diachrony, orthography, universals,
sign language phonology, the phonology/syntax interface,
systems of intonation and stress, and many others besides.
However, numerous bibliographic references are supplied at
the end of the chapter, and readers may wish to consult
these other works.

\section{Early Generative Phonology}

Some key concepts of phonology are best introduced by way
of simple examples involving real data.  We begin with some
data from Russian in (\ref{ex:russian1}).  The example shows
some nouns, in nominative and dative cases, transcribed using
the International Phonetic Alphabet.  Note that \ling{x} is the symbol for
a voiceless velar fricative (e.g.\ the \ling{ch} of Scottish \ling{loch}).

\begin{examples}
\item\label{ex:russian1}
\begin{tabular}[t]{lll}
\colname{Nominative} & \colname{Dative} & \colname{Gloss}\\
xlep & xlebu & `bread' \\
grop & grobu & `coffin' \\
sat & sadu & `garden' \\
prut & prudu & `pond'\\
rok & rogu & `horn' \\
ras & razu & `time'
\end{tabular}
\end{examples}

Observe that the dative form involves suffixation of \ling{-u}, and
a change to the final consonant of the nominative form.
In (\ref{ex:russian1}) we see four changes: \ling{p} becomes \ling{b},
\ling{t} becomes \ling{d}, \ling{k} becomes \ling{g},
and \ling{s} becomes \ling{z}.

Where they differ is in their \term{voicing}; for example,
\ling{b} is a \term{voiced} version of \ling{p}, since \ling{b}
involves periodic vibration of the vocal folds, while \ling{p} does not.
The same applies to the other pairs of sounds.  Now we see
that the changes we observed in (\ref{ex:russian1}) are actually
quite systematic.  Such systematic patterns are called
\term{alternations}, and this particular one is known as a
\term{voicing alternation}.  We can formulate this alternation
using a \term{phonological rule} as follows:

\begin{examples}
\item\label{ex:spe1}
\[
  \left[
  \begin{array}{c}
    \mbox{C}\\
    -\mbox{voiced}
  \end{array}
  \right]
  \rightarrow
  \left[
    +\mbox{voiced}
  \right]
  {\Huge /}
  \underline{\hspace*{3ex}} \mbox{V}
\]
{\it
A consonant becomes voiced in the presence of a following vowel
}
\end{examples}

Rule (\ref{ex:spe1}) uses the format of early generative phonology.
In this notation, C represents any consonant and V represents any vowel.
The rule says that, if a voiceless consonant appears in the
\term{phonological environment} `\underline{\hspace*{3ex}} V'
(i.e. preceding a vowel), then the consonant becomes voiced.
By default, vowels have the feature
\(\left[+\mbox{voiced}\right]\), and so can make the observation
that the consonant \term{assimilates} the voicing feature of the following
vowel.

One way to see if our analysis generalises is to check for
any nominative forms that end in a voiced consonant.  We
expect this consonant to stay the same in the dative form.
However, it turns out that we do not find any nominative
forms ending in a voiced consonant.  Rather, we see the
pattern in example (\ref{ex:russian2}).
(Note that \v{c} is an alternative symbol for IPA \textteshlig).

\begin{examples}
\item\label{ex:russian2}
\begin{tabular}[t]{lll}
\colname{Nominative} & \colname{Dative} & \colname{Gloss}\\
\v{c}erep & \v{c}erepu & `skull'\\
xolop & xolopu & `bondman' \\
trup & trupu & `corpse' \\
cvet & cvetu & `colour'\\
les & lesu & `forest' \\
porok & poroku & `vice'
\end{tabular}
\end{examples}

For these words, the voiceless consonants of the nominative
form are unchanged in the dative form, contrary to our
rule (\ref{ex:spe1}).  These cannot be treated as exceptions,
since this second pattern is quite pervasive.  A solution
is to construct an artificial form which is the dative wordform
minus the \ling{-u} suffix.  We will call this the \term{underlying form}
of the word.  Example (\ref{ex:russian3}) illustrates this for
two cases:

\begin{examples}
\item\label{ex:russian3}
\begin{tabular}[t]{llll}
\colname{Underlying} &
\colname{Nominative} & \colname{Dative} & \colname{Gloss}\\
prud & prut & prudu & `pond'\\
cvet & cvet & cvetu & `colour'
\end{tabular}
\end{examples}

Now we can account for the dative form simply by suffixing the
\ling{-u}.  We account for the nominative form with the following
\term{devoicing rule}:

\begin{examples}
\item\label{ex:spe2}
\[
  \left[
  \begin{array}{c}
    \mbox{C}\\
    +\mbox{voiced}
  \end{array}
  \right]
  \rightarrow
  \left[
    -\mbox{voiced}
  \right]
  {\Huge /}
  \underline{\hspace*{3ex}} \#
\]
{\it
A consonant becomes devoiced word-finally
}
\end{examples}

This rule states that a voiced consonant is devoiced
(i.e. [+voiced] becomes [--voiced]) if the consonant is followed
by a word boundary (symbolised by \#).
It solves a problem with rule~\ref{ex:spe1} which only accounts for half of
the data.  Rule~\ref{ex:spe2} is called
a \term{neutralisation} rule, because the \term{voicing contrast}
of the underlying form is removed in the nominative form.
Now the analysis
accounts for all the nominative and dative forms.
Typically, rules like (\ref{ex:spe2}) can simultaneously
employ several of the distinctive features from (\ref{ex:distinctive}).

Observe that our analysis involves a certain degree of \term{abstractness}.
We have constructed a new \term{level of representation} and drawn
inferences about the \term{underlying forms} by inspecting the
observed \term{surface forms}.

To conclude the development so far, we have seen a simple
kind of \term{phonological representation} (namely sequences
of alphabetic symbols, where each stands for a bundle of
distinctive features), a distinction between levels of
representation, and rules which account for the relationship
between the representations on various levels.  One way or another,
most of phonology is concerned about these three things:
representations, levels, and rules.

Finally, let us consider the plural forms shown in example
(\ref{ex:russian4}).  The plural morpheme is either \ling{-a} or \ling{-y}.

\begin{examples}
\item\label{ex:russian4}
\begin{tabular}[t]{llll}
\colname{Singular} & \colname{Plural} & \colname{Gloss}\\
xlep & xleba & `bread' \\
grop & groby & `coffin' \\
\v{c}erep & \v{c}erepa & `skull'\\
xolop & xolopy & `bondman' \\
trup & trupy & `corpse' \\
sat & sady & `garden' \\
prut & prudy & `pond'\\
cvet & cveta & `colour'\\
ras & razy & `time' \\
les & lesa & `forest' \\
rok & roga & `horn' \\
porok & poroky & `vice'
\end{tabular}
\end{examples}

The phonological environment of the suffix provides us with no way of
predicting which allomorph is chosen.  One solution would be to
enrich the underlying form once more (for example, we could include
the plural suffix in the underlying form, and then have rules to
delete it in all cases but the plural).  A better approach in this
case is to distinguish two \term{morphological classes}, one for
nouns taking the \ling{-y} plural, and one for nouns taking the
\ling{-a} plural.
This information would then be an idiosyncratic property of each lexical
item, and a morphological rule would be responsible for the choice between
the \ling{-y} and \ling{-a} \term{allomorphs}.
A full account of this data, then, must involve phonological, morphological
and lexical modules of a grammar.

As another example, let us consider the vowels of Turkish.
These vowels are tabulated below, along
with a decomposition into distinctive features: [high], [back] and [round].
The features [high] and [back] relate to the position of the tongue
body in the oral cavity.  The feature [round] relates to the rounding
of the lips, as in the English \ling{w} sound.\footnote{Note
that there is a distinction made in the Turkish
alphabet between the dotted \ling{i} and the dotless \ling{\i}.
This \ling{\i} is a high, back, unrounded
vowel that does not occur in English.}

\begin{examples}
\item\label{ex:turkish-vowels}
\begin{tabular}[t]{l|llllllll}
& u & o & \"u & \"o & \i & a & i & e \\ \hline
high & +  & -- & +  & -- & +  & -- & +  & -- \\
back & +  & +  & -- & -- & +  & +  & -- & -- \\
round& +  & +  & +  & +  & -- & -- & -- & -- \\
\end{tabular}
\end{examples}
Consider the following Turkish words, paying
particular attention to the four versions of the possessive
suffix.  Note that similar data are discussed in chapter 2.

\begin{examples}
\item\label{ex:turkish-words1}
\begin{tabular}[t]{llll}
  ip & `rope' & ipin & `rope's' \\
  k\i z & `girl' & k\i z\i n & `girl's' \\
  y\"uz & `face' & y\"uz\"un & `face's' \\
  pul & `stamp' & pulun & `stamp's' \\
  el & `hand' & elin & `hand's' \\
  \c{c}an & `bell' & \c{c}an\i n & `bell's' \\
  k\"oy & `village' & k\"oy\"un & `village's' \\
  son & `end' & sonun & `end's'
\end{tabular}
\end{examples}

The possessive suffix has the forms \ling{in}, \ling{\i n},
\ling{\"un} and \ling{un}.  In terms of the distinctive
feature chart in (\ref{ex:turkish-vowels}), we can observe
that the suffix vowel is always [+high].  The other features
of the suffix vowel are copied from the stem vowel.
This copying is called \term{vowel harmony}.
Let us see how this behaviour can be expressed using a
phonological rule.  To do this, we assume that the vowel
of the possessive affix
is only specified as [+high] and is \term{underspecified}
for its other features.  In the following rule,
\ling{C} denotes any consonant, and the Greek letter variables
range over the + and -- values of the feature.

\begin{examples}
\item\label{ex:SPE-Turkish}
\[
\left[
  \begin{array}{l}
     V \\ \mbox{+high}
  \end{array}
\right] \longrightarrow
\left[
  \begin{array}{l}
     \alpha\mbox{back} \\ \beta\mbox{round}
  \end{array}
\right] \mbox{\Huge /}
\left[
  \begin{array}{l}
     \alpha\mbox{back} \\ \beta\mbox{round}
  \end{array}
\right] C^* \mbox{\underline{\makebox[3ex]{}}}
\]
{\it
A high vowel assimilates to the backness and rounding of the preceding vowel
}
\end{examples}
So long as the stem vowel is specified for the
properties [high] and [back], this rule will make sure that they
are copied onto the affix vowel.  However, there is nothing
in the rule formalism to stop the variables being used in
inappropriate ways (e.g. $\alpha$ back $\rightarrow$ $\alpha$ round).
So we can see that the rule formalism does not permit us to express
the notion that certain features are \term{shared}
by more than one segment.  Instead, we would like to
be able to represent the sharing explicitly, as follows,
where $\pm$H abbreviates [$\pm$high], an underspecified vowel position:

\begin{examples}
\item\label{ex:ap1}\hfil\\

\begin{minipage}[t]{0.4\textwidth}
{\setlength{\unitlength}{1mm}
\begin{picture}(50,35)(5,0)
\put(10,30){\makebox(0,0)[c]{\strut \c{c}}}
\put(20,30){\makebox(0,0)[c]{\strut --H}}
\put(30,30){\makebox(0,0)[c]{\strut n}}
\put(40,30){\makebox(0,0)[c]{\strut +H}}
\put(50,30){\makebox(0,0)[c]{\strut n}}
\put(30,10){\makebox(0,0)[c]{\strut +back}}
\put(30,5){\makebox(0,0)[c]{\strut --round}}
\put(30,15){\line(-1,1){10}}
\put(30,15){\line(1,1){10}}
\end{picture}
}
\end{minipage}
\hfil
\begin{minipage}[t]{0.4\textwidth}
{\setlength{\unitlength}{1mm}
\begin{picture}(50,35)(5,0)
\put(10,30){\makebox(0,0)[c]{\strut k}}
\put(20,30){\makebox(0,0)[c]{\strut +H}}
\put(30,30){\makebox(0,0)[c]{\strut y}}
\put(40,30){\makebox(0,0)[c]{\strut +H}}
\put(50,30){\makebox(0,0)[c]{\strut n}}
\put(30,10){\makebox(0,0)[c]{\strut --back}}
\put(30,5){\makebox(0,0)[c]{\strut +round}}
\put(30,15){\line(-1,1){10}}
\put(30,15){\line(1,1){10}}
\end{picture}
}
\end{minipage}
\end{examples}
The lines of this diagram indicate that the backness and roundness
properties are shared by both vowels in a word.  A single
vowel property (or type) is manifested
on two separate vowels (tokens).

Entities like [+back,--round] that function over extended regions
are often referred to as \term{prosodies}, and this kind of picture is
sometimes called a \term{non-linear} representation.  Many
phonological models use non-linear representations of one sort
or another.  Here we shall consider one particular model,
namely \term{autosegmental phonology}, since it is the most widely
used non-linear model.  The term comes from `autonomous + segment',
and refers to the autonomous nature of segments (or certain groups of
features) once they have been liberated from one-dimensional strings.

\section{Autosegmental Phonology}

In autosegmental phonology, diagrams like those we saw above are known
as \term{charts}.  A chart consists of two or more \term{tiers}, along with
some \term{association lines} drawn between the autosegments on those
tiers.  The \term{no-crossing constraint} is a stipulation that
association lines are not allowed to cross, ensuring that association
lines can be interpreted as asserting some kind of temporal overlap or
inclusion.  \term{Autosegmental rules} are procedures
for converting one representation into another, by adding or removing
association lines and autosegments.  A rule for Turkish vowel harmony
is shown below on the left in (\ref{ex:harmony}),
where \ling{V} denotes any vowel, and the dashed line
indicates that a new association is created.  This rule applies to the
representation in the middle, to yield the one on the right.

\begin{examples}
\item\label{ex:harmony}\hfil\\

{\setlength{\unitlength}{1mm}
\begin{minipage}[t]{0.15\textwidth}
\begin{picture}(25,35)(5,0)
\put(10,30){\makebox(0,0)[c]{\strut V}}
\put(15,30){\makebox(0,0)[c]{\strut C$^*$}}
\put(20,30){\makebox(0,0)[c]{\strut V}}
\put(10,10){\makebox(0,0)[c]{\strut +back}}
\put(10,5){\makebox(0,0)[c]{\strut --round}}
\put(10,15){\line(0,1){10}}
\dashline{3}(10,15)(20,25)
\end{picture}
\end{minipage}}
\hfil
{\setlength{\unitlength}{1mm}
\begin{minipage}[t]{0.35\textwidth}
\begin{picture}(50,35)(5,0)
\put(10,30){\makebox(0,0)[c]{\strut \c{c}}}
\put(20,30){\makebox(0,0)[c]{\strut --H}}
\put(30,30){\makebox(0,0)[c]{\strut n}}
\put(40,30){\makebox(0,0)[c]{\strut +H}}
\put(50,30){\makebox(0,0)[c]{\strut n}}
\put(20,10){\makebox(0,0)[c]{\strut +back}}
\put(20,5){\makebox(0,0)[c]{\strut --round}}
\put(20,15){\line(0,1){10}}
\end{picture}
\end{minipage}}
\hfil
{\setlength{\unitlength}{1mm}
\begin{minipage}[t]{0.35\textwidth}
\begin{picture}(50,35)(5,0)
\put(10,30){\makebox(0,0)[c]{\strut \c{c}}}
\put(20,30){\makebox(0,0)[c]{\strut --H}}
\put(30,30){\makebox(0,0)[c]{\strut n}}
\put(40,30){\makebox(0,0)[c]{\strut +H}}
\put(50,30){\makebox(0,0)[c]{\strut n}}
\put(30,10){\makebox(0,0)[c]{\strut +back}}
\put(30,5){\makebox(0,0)[c]{\strut --round}}
\put(30,15){\line(-1,1){10}}
\put(30,15){\line(1,1){10}}
\end{picture}
\end{minipage}}
\end{examples}

In order to fully appreciate the power of autosegmental phonology,
we will use it to analyse some data from an African tone language.
Consider the data in Table~\ref{tab:chakosi}.  Twelve nouns are listed
down the left side, and the isolation form and five contextual forms
are provided across the table.  The line segments indicate voice pitch
(the fundamental frequency of the voice);
dotted lines are for the syllables of the context words,
and full lines are for
the syllables of the target word, as it is pronounced in this context.
At first glace this data seems bewildering in its complexity.
However, we will see how autosegmental analysis reveals the simple
underlying structure of the data.
\def\slot{\underline{\hspace*{5mm}}}
\begin{table}[t]
{\small\setlength{\tabcolsep}{0.5\tabcolsep}
\begin{tabular}[t]{l|l|l|l|l|l|l}
\colname{}
& A.
& B.
& C.
& D.
& E.
& F.
\\

\colname{Wordform}
& \colname{\slot}
& \colname{i \slot}
& \colname{am \textg oro \slot}
& \colname{\slot\ k\~u}
& \colname{am \slot\ wo d\textopeno}
& \colname{jiine \slot\ ni}
\\

& {\scriptsize isolation}
& {\scriptsize `his ...'}
& {\scriptsize `your (pl)}
& {\scriptsize `one ...'}
& {\scriptsize `your (pl) ...'}
& {\scriptsize `that ...'}
\\

& 
& 
& {\scriptsize brother's ...'}
& 
& {\scriptsize is there'}
& 
\\ \hline

1. {\bf b\textscripta k\textscripta} `tree'
& \l{2}{2} \l{3}{3}
& \d{3}{3} \l{3}{3} \l{2}{2}
& \d{3}{1} \d{1}{1} \d{1}{1} \l{1}{1} \l{2}{2}
& \l{2}{2} \l{3}{3} \d{1}{1}
& \d{3}{1} \l{1}{1} \l{2}{2} \d{2}{2} \d{0}{0}
& \d{2}{2} \d{3}{3} \l{3}{3} \l{2}{2} \d{0}{0}
\\

2. {\bf s\textscripta k\textscripta} `comb'
& \l{2}{2} \l{3}{1}
& \d{3}{3} \l{3}{3} \l{2}{2}
& \d{3}{1} \d{1}{1} \d{1}{1} \l{1}{1} \l{2}{0}
& \l{2}{2} \l{3}{3} \d{1}{1}
& \d{3}{1} \l{1}{1} \l{2}{2} \d{2}{2} \d{0}{0}
& \d{2}{2} \d{3}{3} \l{3}{3} \l{2}{2} \d{0}{0}
\\

3. {\bf buri} `duck'
& \l{3}{3} \l{1}{1}
& \d{3}{3} \l{3}{3} \l{1}{1}
& \d{3}{1} \d{1}{1} \d{1}{1} \l{2}{2} \l{0}{0}
& \l{3}{3} \l{3}{3} \d{1}{1}
& 
& \d{2}{2} \d{3}{3} \l{3}{3} \l{3}{3} \d{0}{0}
\\

4. {\bf siri} `goat'
& \l{3}{3} \l{3}{1}
& \d{3}{3} \l{3}{3} \l{3}{1}
& \d{3}{1} \d{1}{1} \d{1}{1} \l{2}{2} \l{2}{0}
& \l{3}{3} \l{3}{3} \d{1}{1}
& \d{3}{1} \l{2}{2} \l{2}{2} \d{2}{2} \d{0}{0}
& \d{2}{2} \d{3}{3} \l{3}{3} \l{3}{3} \d{0}{0}
\\

5. {\bf \textg\textscripta do} `bed'
& \l{2}{2} \l{2}{2}
& \d{3}{3} \l{3}{3} \l{3}{3}
& \d{3}{1} \d{1}{1} \d{1}{1} \l{2}{2} \l{2}{2}
& \l{3}{3} \l{3}{3} \d{1}{1}
& 
& \d{2}{2} \d{3}{3} \l{3}{3} \l{3}{3} \d{0}{0}
\\

6. {\bf \textg\textopeno r\textopeno} `brother'
& \l{1}{1} \l{1}{1}
& \d{3}{3} \l{3}{3} \l{1}{1}
& 
& \l{1}{1} \l{1}{1} \d{1}{1}
& \d{3}{1} \l{1}{1} \l{1}{1} \d{2}{2} \d{0}{0}
& 
\\ \hline

7. {\bf c\textscripta} `dog'
& \l{1}{3}
& \d{3}{3} \l{1}{1}
& \d{3}{1} \d{1}{1} \d{1}{1} \l{1}{3}
& \l{1}{1} \d{1}{1}
& \d{3}{1} \l{1}{1} \d{2}{2} \d{0}{0}
& \d{2}{2} \d{3}{3} \l{3}{3} \d{0}{0}
\\

8. {\bf ni} `mother'
& \l{1}{1}
& \d{1}{1} \l{1}{1}
& \d{3}{1} \d{1}{1} \d{1}{1} \l{2}{2}
& \l{3}{3} \d{1}{1}
& 
& \d{2}{2} \d{3}{3} \l{3}{3} \d{0}{0}
\\ \hline

9. {\bf j\textopeno k\textopeno r\textopeno} `chain'
& \l{2}{2} \l{2}{2} \l{3}{3}
& \d{3}{3} \l{3}{3} \l{1}{1} \l{2}{2}
& \d{3}{1} \d{1}{1} \d{1}{1} \l{1}{1} \l{1}{1} \l{2}{2}
& \l{2}{2} \l{2}{2} \l{3}{3} \d{1}{1}
& \d{3}{1} \l{1}{1} \l{1}{1} \l{2}{2} \d{2}{2} \d{0}{0}
& \d{2}{2} \d{3}{3} \l{3}{3} \l{1}{1} \l{2}{2} \d{0}{0}
\\

10. {\bf tokoro} `window'
& \l{3}{3} \l{3}{3} \l{3}{3}
& \d{3}{3} \l{3}{3} \l{3}{3} \l{3}{3}
& \d{3}{1} \d{1}{1} \d{1}{1} \l{2}{2} \l{2}{2} \l{2}{2}
& \l{3}{3} \l{3}{3} \l{3}{3} \d{1}{1}
& \d{3}{1} \l{2}{2} \l{2}{2} \l{2}{2} \d{2}{2} \d{0}{0}
& \d{2}{2} \d{3}{3} \l{3}{3} \l{3}{3} \l{3}{3} \d{1}{1}
\\

11. {\bf bul\textscripta li} `iron'
& \l{2}{2} \l{3}{3} \l{1}{1}
& \d{3}{3} \l{3}{3} \l{2}{2} \l{0}{0}
& \d{3}{1} \d{1}{1} \d{1}{1} \l{1}{1} \l{2}{2} \l{0}{0}
& \l{2}{2} \l{3}{3} \l{3}{3} \d{1}{1}
& \d{4}{2} \l{2}{2} \l{3}{3} \l{3}{3} \d{2}{2} \d{0}{0}
& \d{2}{2} \d{3}{3} \l{3}{3} \l{2}{2} \l{2}{2} \d{0}{0}
\\

12. {\bf misini} `needle'
& \l{3}{3} \l{3}{3} \l{1}{1}
& \d{3}{3} \l{3}{3} \l{3}{3} \l{1}{1}
& \d{3}{1} \d{1}{1} \d{1}{1} \l{2}{2} \l{2}{2} \l{0}{0}
& \l{3}{3} \l{3}{3} \l{3}{3} \d{1}{1}
& \d{3}{1} \l{3}{3} \l{3}{3} \l{3}{3} \d{2}{2} \d{0}{0}
& \d{2}{2} \d{3}{3} \l{3}{3} \l{3}{3} \l{3}{3} \d{1}{1}

\end{tabular}}
\caption{Tone Data from Chakosi (Ghana)}\label{tab:chakosi}
\end{table}

Looking across the table, observe that the contextual forms of
a given noun are quite variable.  For example
\ling{bul\textscripta li} appears as
\mbox{\l{2}{2} \l{3}{3} \l{1}{1}},
\mbox{\l{3}{3} \l{2}{2} \l{0}{0}},
\mbox{\l{2}{2} \l{3}{3} \l{3}{3}}, and
\mbox{\l{3}{3} \l{2}{2} \l{2}{2}}.

We could begin the analysis by identifying all the levels (here there are
five), assigning a name or number to each, and looking for patterns.
However, this approach does not capture the relative nature of tone, where
\mbox{\l{1}{1} \l{2}{2} \l{0}{0}} is not distinguished from
\mbox{\l{3}{3} \l{4}{4} \l{2}{2}}.  Instead, our approach just has to be
sensitive to \emph{differences} between adjacent tones.  So these distinct tone
sequences could be represented identically as $+1$, $-2$, since we go
up a small amount from the first to the second tone ($+1$), and then
down a larger amount ($-2$).
In autosegmental analysis, we treat \term{contour tones} as being made up of
two or more \term{level tones} compressed into the space of a single syllable.
Therefore, we can treat \mbox{\l{1}{1} \l{2}{0}} as another instance
of $+1$, $-2$.
Given our autosegmental perspective, a sequence of two or more identical
tones corresponds
to a single spread tone.  This means that
we can collapse sequences of like tones
to a single tone.\footnote{
  This assumption cannot be maintained in more
  sophisticated approaches involving lexical and prosodic domains.
  However, it is a very useful simplifying assumption for the
  purposes of this presentation.
}
When we retranscribe our data in this way, some interesting
patterns emerge.

First, by observing the raw frequency of these
intertone intervals, we see that $-2$ and $+1$ are by
far the most common, occurring 63 and 39 times respectively.
A $-1$ difference occurs 8 times, while a $+2$ difference is very
rare (only occurring 3 times, and only in phrase-final contour tones).
This patterning is characteristic of a \term{terrace tone language}.
In analysing such a language, phonologists typically propose an inventory of
just two tones, H (high) and L (low), where these might be represented
featurally as [$\pm$hi].  In such a model, the tone sequence
HL corresponds to \mbox{\l{3}{3} \l{1}{1}}, a pitch difference of $-2$.

In terrace tone languages, an H tone does not achieve its former level
after an L tone, so HLH is \term{phonetically realized} as
\mbox{\l{3}{3} \l{1}{1} \l{2}{2}},
(instead of \mbox{\l{3}{3} \l{1}{1} \l{3}{3}}).
This kind of H-lowering is called \term{automatic downstep}.
A pitch difference of $+1$ corresponds to an LH tone sequence.
With this model, we already account for the prevalence of
the $-2$ and $+1$ intervals.  What about $-1$ and $+2$?

As we will see later, the $-1$ difference arises when
the middle tone of \mbox{\l{3}{3} \l{1}{1} \l{2}{2}} (HLH) is deleted,
leaving just \mbox{\l{3}{3} \l{2}{2}}.  In this situation we
write H!H, where the exclamation mark indicates the lowering of
the following H due to a deleted (or \term{floating} low tone).
This kind of H-lowering is called
\term{conditioned downstep}.  The rare $+2$ difference only occurs
for an LH contour; we can assume that
automatic downstep only applies when a LH sequence is linked
to two separate syllables (\mbox{\l{1}{1} \l{2}{2}}) and not when
the sequence is linked to a single syllable (\mbox{\l{1}{3}}).

To summarise these conventions, we associate the pitch differences
to tone sequences as shown in (\ref{ex:simple-tone}).  Syllable
boundaries are marked with a dot.

\begin{examples}
\item\label{ex:simple-tone}
\begin{tabular}[t]{rrrrr}
\colname{Interval}  & $-2$  & $-1$    & $+1$   & $+2$  \\
\colname{Pitches}   & \l{2}{2} \l{0}{0}
                    & \l{2}{2} \l{1}{1}
                    & \l{1}{1} \l{2}{2}
                    & \l{0}{2} \\
\colname{Tones}     & H.L & H.!H  & L.H & LH
\end{tabular}
\end{examples}

Now we are in a position to provide tonal transcriptions
for the forms in Table~\ref{tab:chakosi}.  Example (\ref{ex:trans})
gives the transcriptions for the forms involving \ling{bul\textscripta li}.
Tones corresponding to the noun are underlined.

\begin{examples}
\item\label{ex:trans}
{\bf Transcriptions of bul\textscripta li `iron'}

\begin{tabular}{llll}
bul\textscripta li
& `iron'
& \l{2}{2} \l{3}{3} \l{1}{1}
& \underline{L.H.L}
\\

i bul\textscripta li
& `his iron'
& \d{3}{3} \l{3}{3} \l{2}{2} \l{0}{0}
& H.\underline{H.!H.L}
\\

am \textg oro bul\textscripta li
& `your (pl) brother's iron'
& \d{3}{1} \d{1}{1} \d{1}{1} \l{1}{1} \l{2}{2} \l{0}{0}
& HL.L.L.\underline{L.H.L}
\\

bul\textscripta li k\~u
& `one iron'
& \l{2}{2} \l{3}{3} \l{3}{3} \d{1}{1}
& \underline{L.H.H}.L
\\

am bul\textscripta li wo d\textopeno
& `your (pl) iron is there'
& \d{4}{2} \l{2}{2} \l{3}{3} \l{3}{3} \d{2}{2} \d{0}{0}
& HL.\underline{L.H.H}.!H.L
\\

jiine bul\textscripta li ni
& `that iron'
& \d{2}{2} \d{3}{3} \l{3}{3} \l{2}{2} \l{2}{2} \d{0}{0}
& L.H.\underline{H.!H.H}.L
\end{tabular}
\end{examples}

Looking down the right hand column of (\ref{ex:trans}) at the
underlined tones, observe again the diversity of \term{surface forms}
corresponding to the single lexical item.  An autosegmental analysis
is able to account for all this variation with a single spreading
rule.

\begin{examples}
\item\label{ex:hts}
{\bf High Tone Spread}\\

{\setlength{\unitlength}{1mm}
\begin{minipage}[t]{0.15\textwidth}
\begin{picture}(25,35)(5,0)
\put(10,30){\makebox(0,0)[c]{\strut $\sigma$}}
\put(20,30){\makebox(0,0)[c]{\strut $\sigma$}}
\put(30,30){\makebox(0,0)[c]{\strut $\sigma$}}
\put(10,10){\makebox(0,0)[c]{\strut H}}
\put(20,10){\makebox(0,0)[c]{\strut L}}
\put(10,15){\line(0,1){10}}
\put(20,15){\line(0,1){10}}
\put(20,20){\makebox(0,0)[c]{\strut =}}
\dashline{3}(10,15)(20,25)
\end{picture}
\end{minipage}}
{\it 
A high tone spreads to the following (non-final) syllable, delinking the low tone
}
\end{examples}

Rule (\ref{ex:hts}) applies to any sequence of three syllables ($\sigma$)
where the first is linked to an H tone and the second is linked to
an L tone.  The rule spreads H to the right, delinking the L.
Crucially, the L itself is not deleted, but remains as a
\term{floating tone}, and continues to influence surface tone
as downstep.  Example (\ref{ex:hts-app}) shows the
application of the H spread rule to forms involving
\ling{bul\textscripta li}.  The first row of autosegmental diagrams
shows the underlying forms, where \ling{bul\textscripta li} is
assigned an LHL \term{tone melody}.  In the second row, we see
the result of applying H spread.  Following standard practice, the
floating low tones are circled.  Where a floating L appears between
two H tones, it gives rise to downstep.  The final assignment of tones
to syllables and the position of the downsteps are shown in the last
row of the table.

\begin{examples}
\item\label{ex:hts-app}

{\setlength{\tabcolsep}{0.9\tabcolsep}
\begin{tabular}[t]{llll}
B. `his iron'
& D. `one iron'
& E. `your (pl) iron'
& F. `that iron'
\\

{\setlength{\unitlength}{.6mm}
\begin{minipage}[t]{0.15\textwidth}
\begin{picture}(45,35)(5,0)
\put(10,30){\makebox(0,0)[c]{\strut i}}
\put(20,30){\makebox(0,0)[c]{\strut bu}}
\put(30,30){\makebox(0,0)[c]{\strut l\textscripta}}
\put(40,30){\makebox(0,0)[c]{\strut li}}
\put(10,10){\makebox(0,0)[c]{\strut H}}
\put(20,10){\makebox(0,0)[c]{\strut L}}
\put(30,10){\makebox(0,0)[c]{\strut H}}
\put(40,10){\makebox(0,0)[c]{\strut L}}
\put(10,15){\line(0,1){10}}
\put(20,15){\line(0,1){10}}
\put(30,15){\line(0,1){10}}
\put(40,15){\line(0,1){10}}
\end{picture}
\end{minipage}}
&
{\setlength{\unitlength}{.6mm}
\begin{minipage}[t]{0.15\textwidth}
\begin{picture}(45,35)(5,0)
\put(10,30){\makebox(0,0)[c]{\strut bu}}
\put(20,30){\makebox(0,0)[c]{\strut l\textscripta}}
\put(30,30){\makebox(0,0)[c]{\strut li}}
\put(40,30){\makebox(0,0)[c]{\strut k\~u}}
\put(10,10){\makebox(0,0)[c]{\strut L}}
\put(20,10){\makebox(0,0)[c]{\strut H}}
\put(30,10){\makebox(0,0)[c]{\strut L}}
\put(40,10){\makebox(0,0)[c]{\strut L}}
\put(10,15){\line(0,1){10}}
\put(20,15){\line(0,1){10}}
\put(30,15){\line(0,1){10}}
\put(40,15){\line(0,1){10}}
\end{picture}
\end{minipage}}
&
{\setlength{\unitlength}{.6mm}
\begin{minipage}[t]{0.15\textwidth}
\begin{picture}(45,35)(5,0)
\put(10,30){\makebox(0,0)[c]{\strut \textscripta m}}
\put(20,30){\makebox(0,0)[c]{\strut bu}}
\put(30,30){\makebox(0,0)[c]{\strut l\textscripta}}
\put(40,30){\makebox(0,0)[c]{\strut li}}
\put(50,30){\makebox(0,0)[c]{\strut wo}}
\put(60,30){\makebox(0,0)[c]{\strut d\textopeno}}
\put( 7,10){\makebox(0,0)[c]{\strut H}}
\put(13,10){\makebox(0,0)[c]{\strut L}}
\put(20,10){\makebox(0,0)[c]{\strut L}}
\put(30,10){\makebox(0,0)[c]{\strut H}}
\put(40,10){\makebox(0,0)[c]{\strut L}}
\put(50,10){\makebox(0,0)[c]{\strut H}}
\put(60,10){\makebox(0,0)[c]{\strut L}}
\put(8,15){\line(1,5){2}}
\put(12,15){\line(-1,5){2}}
\put(20,15){\line(0,1){10}}
\put(30,15){\line(0,1){10}}
\put(40,15){\line(0,1){10}}
\put(50,15){\line(0,1){10}}
\put(60,15){\line(0,1){10}}
\end{picture}
\end{minipage}}
&
{\setlength{\unitlength}{.6mm}
\begin{minipage}[t]{0.15\textwidth}
\begin{picture}(45,35)(5,0)
\put(10,30){\makebox(0,0)[c]{\strut jii}}
\put(20,30){\makebox(0,0)[c]{\strut ni}}
\put(30,30){\makebox(0,0)[c]{\strut bu}}
\put(40,30){\makebox(0,0)[c]{\strut l\textscripta}}
\put(50,30){\makebox(0,0)[c]{\strut li}}
\put(60,30){\makebox(0,0)[c]{\strut ni}}
\put(10,10){\makebox(0,0)[c]{\strut L}}
\put(20,10){\makebox(0,0)[c]{\strut H}}
\put(30,10){\makebox(0,0)[c]{\strut L}}
\put(40,10){\makebox(0,0)[c]{\strut H}}
\put(50,10){\makebox(0,0)[c]{\strut L}}
\put(60,10){\makebox(0,0)[c]{\strut L}}
\put(10,15){\line(0,1){10}}
\put(20,15){\line(0,1){10}}
\put(30,15){\line(0,1){10}}
\put(40,15){\line(0,1){10}}
\put(50,15){\line(0,1){10}}
\put(60,15){\line(0,1){10}}
\end{picture}
\end{minipage}}
\\

{\setlength{\unitlength}{.6mm}
\begin{minipage}[t]{0.15\textwidth}
\begin{picture}(45,35)(5,0)
\put(10,30){\makebox(0,0)[c]{\strut i}}
\put(20,30){\makebox(0,0)[c]{\strut bu}}
\put(30,30){\makebox(0,0)[c]{\strut l\textscripta}}
\put(40,30){\makebox(0,0)[c]{\strut li}}
\put(10,10){\makebox(0,0)[c]{\strut H}}
\put(20,10){\makebox(0,0)[c]{\strut \ovalbox{L}}}
\put(30,10){\makebox(0,0)[c]{\strut H}}
\put(40,10){\makebox(0,0)[c]{\strut L}}
\put(10,15){\line(0,1){10}}
\put(10,15){\line(1,1){10}}
\put(30,15){\line(0,1){10}}
\put(40,15){\line(0,1){10}}
\end{picture}
\end{minipage}}
&
{\setlength{\unitlength}{.6mm}
\begin{minipage}[t]{0.15\textwidth}
\begin{picture}(45,35)(5,0)
\put(10,30){\makebox(0,0)[c]{\strut bu}}
\put(20,30){\makebox(0,0)[c]{\strut l\textscripta}}
\put(30,30){\makebox(0,0)[c]{\strut li}}
\put(40,30){\makebox(0,0)[c]{\strut k\~u}}
\put(10,10){\makebox(0,0)[c]{\strut L}}
\put(20,10){\makebox(0,0)[c]{\strut H}}
\put(30,10){\makebox(0,0)[c]{\strut \ovalbox{L}}}
\put(40,10){\makebox(0,0)[c]{\strut L}}
\put(10,15){\line(0,1){10}}
\put(20,15){\line(0,1){10}}
\put(20,15){\line(1,1){10}}
\put(40,15){\line(0,1){10}}
\end{picture}
\end{minipage}}
&
{\setlength{\unitlength}{.6mm}
\begin{minipage}[t]{0.15\textwidth}
\begin{picture}(45,35)(5,0)
\put(10,30){\makebox(0,0)[c]{\strut \textscripta m}}
\put(20,30){\makebox(0,0)[c]{\strut bu}}
\put(30,30){\makebox(0,0)[c]{\strut l\textscripta}}
\put(40,30){\makebox(0,0)[c]{\strut li}}
\put(50,30){\makebox(0,0)[c]{\strut wo}}
\put(60,30){\makebox(0,0)[c]{\strut d\textopeno}}
\put( 7,10){\makebox(0,0)[c]{\strut H}}
\put(13,10){\makebox(0,0)[c]{\strut L}}
\put(20,10){\makebox(0,0)[c]{\strut L}}
\put(30,10){\makebox(0,0)[c]{\strut H}}
\put(40,10){\makebox(0,0)[c]{\strut \ovalbox{L}}}
\put(50,10){\makebox(0,0)[c]{\strut H}}
\put(60,10){\makebox(0,0)[c]{\strut L}}
\put(8,15){\line(1,5){2}}
\put(12,15){\line(-1,5){2}}
\put(20,15){\line(0,1){10}}
\put(30,15){\line(0,1){10}}
\put(30,15){\line(1,1){10}}
\put(50,15){\line(0,1){10}}
\put(60,15){\line(0,1){10}}
\end{picture}
\end{minipage}}
&
{\setlength{\unitlength}{.6mm}
\begin{minipage}[t]{0.15\textwidth}
\begin{picture}(45,35)(5,0)
\put(10,30){\makebox(0,0)[c]{\strut jii}}
\put(20,30){\makebox(0,0)[c]{\strut ni}}
\put(30,30){\makebox(0,0)[c]{\strut bu}}
\put(40,30){\makebox(0,0)[c]{\strut l\textscripta}}
\put(50,30){\makebox(0,0)[c]{\strut li}}
\put(60,30){\makebox(0,0)[c]{\strut ni}}
\put(10,10){\makebox(0,0)[c]{\strut L}}
\put(20,10){\makebox(0,0)[c]{\strut H}}
\put(30,10){\makebox(0,0)[c]{\strut \ovalbox{L}}}
\put(40,10){\makebox(0,0)[c]{\strut H}}
\put(50,10){\makebox(0,0)[c]{\strut \ovalbox{L}}}
\put(60,10){\makebox(0,0)[c]{\strut L}}
\put(10,15){\line(0,1){10}}
\put(20,15){\line(0,1){10}}
\put(20,15){\line(1,1){10}}
\put(40,15){\line(0,1){10}}
\put(40,15){\line(1,1){10}}
\put(60,15){\line(0,1){10}}
\end{picture}
\end{minipage}}
\\

{\setlength{\tabcolsep}{0.5\tabcolsep}
\begin{tabular}{llll}
i & bu & l\textscripta & li \\
H & H  & !H & L \\
\d{3}{3} &\l{3}{3} &\l{2}{2} &\l{0}{0}
\end{tabular}}
&
{\setlength{\tabcolsep}{0.5\tabcolsep}
\begin{tabular}{llll}
bu & l\textscripta & li & k\~u \\
L & H  & H & L \\
\l{2}{2} &\l{3}{3} &\l{3}{3} &\d{1}{1}
\end{tabular}}
&
{\setlength{\tabcolsep}{0.5\tabcolsep}
\begin{tabular}{llllll}
\textscripta m & bu & l\textscripta & li & wo & d\textopeno \\
HL & L  & H & H & !H & L\\
\d{4}{2} &\l{2}{2} &\l{3}{3} &\l{3}{3} &\d{2}{2} &\d{0}{0}
\end{tabular}}
&
{\setlength{\tabcolsep}{0.5\tabcolsep}
\begin{tabular}{llllll}
jii & ni & bu & l\textscripta & li & ni \\
L & H & H & !H & H & L \\
\d{2}{2} & \d{3}{3} & \l{3}{3} & \l{2}{2} & \l{2}{2} & \d{0}{0}
\end{tabular}}
\\

\end{tabular}}
\end{examples}

Example (\ref{ex:hts-app}) shows the power of autosegmental phonology --
together with suitable underlying forms and
appropriate principles of phonetic interpretation --
in analysing complex patterns with simple rules.
Space precludes a full analysis of the data; interested readers
can try hypothesising underlying forms for the other words, along
with new rules, to account for the rest of the data in
Table~\ref{tab:chakosi}.

The preceding discussion of segmental and autosegmental phonology
highlights the multi-linear organisation of phonological representations,
which derives from the temporal nature of the speech stream.
Phonological representations are also organised hierarchically.
We already know that phonological information comprises words,
and words, phrases.  This is one kind of hierarchical organisation
of phonological information.  But phonological analysis has also
demonstrated the need for other kinds of hierarchy, such as
the \term{prosodic hierarchy},
which builds structure involving syllables, feet
and intonational phrases above the segment level, and
\term{feature geometry}, which involves hierarchical organisation beneath
the level of the segment.
Phonological rules and constraints can refer to the prosodic hierarchy in
order to account for the observed \term{distribution} of phonological
information across the linear sequence of segments.  Feature geometry
serves the dual purpose of accounting for the inventory of contrastive
sounds available to a language, and for the alternations we can observe.
Here we will consider just one level of phonological hierarchy, namely
the syllable.

\section{Syllable Structure}

Syllables are a fundamental organisational unit in phonology.
In many languages, phonological alternations are sensitive to syllable
structure.  For example, \ling{t} has several \term{allophones} in
English, and the choice of allophone depends on phonological context.
For example, in many English dialects, \ling{t} is pronounced as
the flap [\textfishhookr] between vowels, as in \ling{water}.
Two other variants are shown in (\ref{ex:t}), where the phonetic
transcription is given in brackets, and syllable boundaries are
marked with a dot.

\begin{examples}
\item\label{ex:t}
\begin{subexamples}
\item
  atlas [\ae t\textsuperscript{\textglotstop}.l\textschwa s]
\item
  cactus [k\ae k.t\textsuperscript{h}\textschwa s]
\end{subexamples}
\end{examples}

Native English syllables cannot begin with \ling{tl}, and so
the \ling{t} of \ling{atlas} is syllabified with the preceding
vowel.  Syllable final \ling{t} is regularly glottalised or unreleased
in English, while syllable initial \ling{t} is regularly aspirated.
Thus we have a natural explanation for the patterning of
these allophones in terms of syllable structure.

Other evidence for the syllable comes from loanwords.  When words are
borrowed into one language from another, they must be adjusted so as
to conform to the legal sound patterns (or \term{phonotactics}) of the
host language.  For example, consider the following borrowings from
English into Dschang, a language of Cameroon \citep{Bird99syl}.

\begin{examples}
\item\label{ex:dschang}
afruwa {\it flower},
akalatusi {\it eucalyptus},
al{\textepsilon}sa {\it razor},
al{\textopeno}ba {\it rubber},
apl{\textepsilon}\textipa{\ng}g{\textepsilon} {\it blanket},
as{\textschwa}kuu {\it school},
c{\textepsilon}{\textepsilon}n {\it chain},
d{\textschwa}{\textschwa}k {\it debt},
kapinda {\it carpenter},
k{\textepsilon}si\textipa{\ng} {\it kitchen},
kuum {\it comb},
laam {\it lamp},
l{\textepsilon}si {\it rice},
luum {\it room},
mbas{\textschwa}ku {\it bicycle},
mbrusi {\it brush},
mb{\textschwa}r{\textschwa}{\textschwa}k {\it brick},
m{\textepsilon}ta {\it mat},
m{\textepsilon}t{\textschwa}rasi {\it mattress},
\textipa{\ng}glasi {\it glass},
{\textltailn}jakasi {\it jackass},
m{\textepsilon}tisi {\it match}
nubatisi {\it rheumatism},
p{\textopeno}k{\textepsilon} {\it pocket}
\textipa{\ng}gal{\textepsilon} {\it garden},
s{\textschwa}sa {\it scissors},
t{\textepsilon}w{\textepsilon}l{\textepsilon} {\it towel},
wasi {\it watch},
zii\textipa{\ng} {\it zinc},
\end{examples}

In Dschang, the \term{syllable canon} is much more restricted
than in English.  Consider the patterning of \ling{t}.  This
segment is illegal in syllable-final position.
In technical language, we would say that alveolars are not
\term{licensed} in the syllable coda.
In m{\textepsilon}ta {\it mat}, a vowel is inserted,
making the \ling{t} into the initial segment of the next syllable.
For d{\textschwa}{\textschwa}k {\it debt}, the place of articulation
of the \ling{t} is changed to velar, making it a legal syllable-final
consonant.
For apl{\textepsilon}\textipa{\ng}g{\textepsilon} {\it blanket}, the
final \ling{t} is deleted.  Many other adjustments
can be seen in (\ref{ex:dschang}), and most of them can be explained
with reference to syllable structure.

A third source of evidence for syllable structure comes from
morphology.  In Ulwa, a Nicaraguan language, the position of
the possessive \term{infix} is sensitive to syllable structure.
The Ulwa syllable canon is (C)V(V|C)(C), and any \term{intervocalic}
consonant (i.e.\ consonant between two vowels) is syllabified with
the following syllable, a universal principle known as
\term{onset maximisation}.
Consider the Ulwa data in (\ref{ex:ulwa}).

\begin{examples}
\item\label{ex:ulwa}
\begin{tabular}[t]{lll|lll}
\colname{Word} & \colname{Possessive} & \colname{Gloss} &
\colname{Word} & \colname{Possessive} & \colname{Gloss} \\ \hline

b\textscripta\textscripta
& b\textscripta\textscripta.{\bf k\textscripta}
& `excrement' &

bi.l\textscripta m
& bi.l\textscripta m.{\bf k\textscripta}
& `fish' \\

dii.muih
& dii.{\bf k\textscripta}.muih
& `snake' &

g\textscripta\textscripta d
& g\textscripta\textscripta d.{\bf k\textscripta}
& `god' \\

ii.bin
& ii.{\bf k\textscripta}.bin
& `heaven' &

ii.li.lih
& ii.{\bf k\textscripta}.li.lih
& `shark' \\

k\textscripta h.m\textscripta
& k\textscripta h.{\bf k\textscripta}.m\textscripta
& `iguana' &

k\textscripta.p\textscripta k
& k\textscripta.p\textscripta k.{\bf k\textscripta}
& `manner' \\

lii.m\textscripta
& lii.{\bf k\textscripta}.m\textscripta
& `lemon' &

mis.tu
& mis.{\bf k\textscripta}.tu
& `cat' \\

on.y\textscripta n
& on.{\bf k\textscripta}.y\textscripta n
& `onion' &

p\textscripta u.m\textscripta k
& p\textscripta u.{\bf k\textscripta}.m\textscripta k
& `tomato' \\

sik.bilh
& sik.{\bf k\textscripta}.bilh
& `horsefly' &

t\textscripta im
& t\textscripta im.{\bf k\textscripta}
& `time' \\

t\textscripta i.t\textscripta i
& t\textscripta i.{\bf k\textscripta}.t\textscripta i
& `grey squirrel' &

uu.m\textscripta k
& uu.{\bf k\textscripta}.m\textscripta k
& `window' \\

w\textscripta i.ku
& w\textscripta i.{\bf k\textscripta}.ku
& `moon, month' &

w\textscripta.s\textscripta.l\textscripta
& w\textscripta.s\textscripta.{\bf k\textscripta}.l\textscripta
& `possum'
\end{tabular}
\end{examples}

Observe that the infix appears at a syllable boundary, and so we
can already state that the infix position is sensitive to syllable
structure.  Any analysis of the infix position must take
\term{syllable weight} into consideration.
Syllables having a single short vowel and no following consonants
are defined to be \term{light}.  (The presence of onset consonants
is irrelevant to syllable weight.)
All other syllables, i.e. those which have two
vowels, or a single long vowel, or a final consonant, are defined
to be \term{heavy};
e.g.\ \ling{kah}, \ling{kaa}, \ling{muih}, \ling{bilh}, \ling{ii},
\ling{on}.
Two common phonological representations for this syllable structure
are the onset-rhyme model, and the moraic model.  Representations
for the syllables just listed are shown in (\ref{ex:syl}).
In these diagrams, $\sigma$ denotes a syllable,
O onset, R rhyme, N nucleus, C coda and $\mu$ \ling{mora}
(the traditional, minimal unit of syllable weight).

\begin{examples}
\item\label{ex:syl}
\begin{subexamples}
\item {\bf The Onset-Rhyme Model of Syllable Structure}\\

\begin{bundle}{$\sigma$}
  \chunk{
    \begin{bundle}{O}\chunk{k}\end{bundle}
  }
  \chunk{
    \begin{bundle}{R}
      \chunk{
        \begin{bundle}{N}\chunk{a}\end{bundle}
      }
    \end{bundle}
  }
\end{bundle}
\hfil
\begin{bundle}{$\sigma$}
  \chunk{
    \begin{bundle}{O}\chunk{k}\end{bundle}
  }
  \chunk{
    \begin{bundle}{R}
      \chunk{\begin{bundle}{N}\chunk{a}\end{bundle}}
      \chunk{\begin{bundle}{C}\chunk{h}\end{bundle}}
    \end{bundle}
  }
\end{bundle}
\hfil
\begin{bundle}{$\sigma$}
  \chunk{
    \begin{bundle}{O}\chunk{k}\end{bundle}
  }
  \chunk{
    \begin{bundle}{R}
      \chunk{
        \begin{bundle}{N}\chunk{a}\chunk{a}\end{bundle}
      }
    \end{bundle}
  }
\end{bundle}
\hfil
\begin{bundle}{$\sigma$}
  \chunk{
    \begin{bundle}{O}\chunk{m}\end{bundle}
  }
  \chunk{
    \begin{bundle}{R}
      \chunk{\begin{bundle}{N}\chunk{u}\chunk{i}\end{bundle}}
      \chunk{\begin{bundle}{C}\chunk{h}\end{bundle}}
    \end{bundle}
  }
\end{bundle}
\hfil
\begin{bundle}{$\sigma$}
  \chunk{
    \begin{bundle}{O}\chunk{b}\end{bundle}
  }
  \chunk{
    \begin{bundle}{R}
      \chunk{\begin{bundle}{N}\chunk{i}\end{bundle}}
      \chunk{\begin{bundle}{C}\chunk{l}\chunk{h}\end{bundle}}
    \end{bundle}
  }
\end{bundle}
\hfil
\begin{bundle}{$\sigma$}
  \chunk{
    \begin{bundle}{R}
      \chunk{\begin{bundle}{N}\chunk{i}\chunk{i}\end{bundle}}
    \end{bundle}
  }
\end{bundle}
\hfil
\begin{bundle}{$\sigma$}
  \chunk{
    \begin{bundle}{R}
      \chunk{\begin{bundle}{N}\chunk{o}\end{bundle}}
      \chunk{\begin{bundle}{C}\chunk{n}\end{bundle}}
    \end{bundle}
  }
\end{bundle}
\vspace{2ex}


\item {\bf The Moraic Model of Syllable Structure}\\

\begin{bundle}{$\sigma$}
  \chunk{k}
  \chunk{
    \begin{bundle}{$\mu$}\chunk{a}\end{bundle}
  }
\end{bundle}
\hfil
\begin{bundle}{$\sigma$}
  \chunk{k}
  \chunk{\begin{bundle}{$\mu$}\chunk{a}\end{bundle}}
  \chunk{\begin{bundle}{$\mu$}\chunk{h}\end{bundle}}
\end{bundle}
\hfil
\begin{bundle}{$\sigma$}
  \chunk{k}
  \chunk{\begin{bundle}{$\mu$}\chunk{a}\end{bundle}}
  \chunk{\begin{bundle}{$\mu$}\chunk{a}\end{bundle}}
\end{bundle}
\hfil
\begin{bundle}{$\sigma$}
  \chunk{m}
  \chunk{\begin{bundle}{$\mu$}\chunk{u}\end{bundle}}
  \chunk{\begin{bundle}{$\mu$}\chunk{i}\chunk{h}\end{bundle}}
\end{bundle}
\hfil
\begin{bundle}{$\sigma$}
  \chunk{b}
  \chunk{\begin{bundle}{$\mu$}\chunk{i}\end{bundle}}
  \chunk{\begin{bundle}{$\mu$}\chunk{l}\chunk{h}\end{bundle}}
\end{bundle}
\hfil
\begin{bundle}{$\sigma$}
  \chunk{\begin{bundle}{$\mu$}\chunk{i}\end{bundle}}
  \chunk{\begin{bundle}{$\mu$}\chunk{i}\end{bundle}}
\end{bundle}
\hfil
\begin{bundle}{$\sigma$}
  \chunk{\begin{bundle}{$\mu$}\chunk{o}\end{bundle}}
  \chunk{\begin{bundle}{$\mu$}\chunk{n}\end{bundle}}
\end{bundle}
\end{subexamples}
\end{examples}

In the onset-rhyme model (\ref{ex:syl}a),
consonants coming before the first vowel are
linked to the onset node, and the rest of the material comes under
the rhyme node.\footnote{Two syllables usually have to agree on the
material in their rhyme constituents in order for them to be
considered rhyming, hence the name.}
A rhyme contains an obligatory nucleus and an optional coda.
In this model, a syllable is said to be heavy if and only if its
rhyme or its nucleus are branching.

In the moraic mode (\ref{ex:syl}b),
any consonants that appear before the first vowel are linked
directly to the syllable node.  The first vowel is linked to
its own mora node (symbolised by $\mu$), and any remaining material is
linked to the second mora node.  A syllable is said to be heavy
if and only if it has more than one mora.

These are just two of several ways that have been proposed for
representing syllable structure.  Now the syllables
constituting a word can now be linked to higher levels of
structure, such as the \ling{foot} and the \ling{prosodic word}.
For now, it is sufficient to know that such higher levels
exist, and that we have a way to
represent the binary distinction of
syllable weight.

Now we can return to the Ulwa data, from example (\ref{ex:ulwa}).
A relatively standard way to account for the infix position is
to stipulate that the first light syllable, if present, is
actually invisible to the rules which assign syllables to
higher levels; such syllables are said to be \term{extra-metrical}.
They are a sort of `upbeat' to the word, and are often associated
with the preceding word in continuous speech.
Given these general principles concerning hierarchical structure,
we can simply state that the Ulwa possessive affix is infixed after the
first syllable.\footnote{
  A better analysis of the Ulwa infixation data involves reference to
\term{metrical feet}, phonological units above the level of the syllable.
This is beyond the scope of the current chapter however.
}

In the foregoing discussion, I hope to have revealed many interesting
issues which are confronted by phonological analysis, without delving
too deeply into the abstract theoretical constructs which phonologists
have proposed.
Theories differ enormously in their organisation of phonological
information and the ways in which they permit this information to be
subjected to rules and constraints, and the way the information is used
in a \term{lexicon} and an overarching \term{grammatical framework}.
Some of these theoretical frameworks include:
lexical phonology, underspecification phonology, government phonology,
declarative phonology, and optimality theory.
For more information about these, please
see \S\ref{sec:reading} for literature references.

\section{Computational phonology}

When phonological information is treated as a string
of atomic symbols, it is immediately amenable to processing
using existing models.  A particularly successful example
is the work on finite state transducers (see chapter 21).  However,
phonologists abandoned linear representations in the 1970s,
and so we will consider some computational models that have been
proposed for multi-linear, hierarchical, phonological representations.
It turns out that these pose some interesting challenges.

Early models of generative phonology, like that of the Sound Pattern of
English (SPE), were sufficiently explicit that they
could be implemented directly.  A necessary first step
in implementing many of the more recent theoretical models is to formalise
them, and to discover the intended semantics of some subtle, graphical
notations.  A practical approach to this problem has been to try to express
phonological information using existing, well-understood computational models.
The principal models are finite state devices and attribute-value matrices.

\subsection{Finite state models of non-linear phonology}

Finite state machines cannot process structured data, only strings, so special
methods are required for these devices to process complex phonological
representations.  All approaches involve a many-to-one mapping from the
parallel layers of representation to a single machine.  There are
essentially three places where this many-to-one mapping can be situated.
The first approach is to employ multi-tape machines \citep{Kay87}.
Each tier is represented as a string, and the set of strings is processed
simultaneously by a single machine.  The second approach
is to map the multiple layers into a single string, and to process that
with a conventional single-tape machine \citep{Kornai95}.  The third approach
is to encode each layer itself as a finite state
machine, and to combine the machines using automaton intersection
\citep{BirdEllison94}.

This work demonstrates how representations can be compiled into a form that
can be directly manipulated by finite state machines.  Independently
of this, we also need to provide a means for phonological
generalisations (such as rules and constraints) to be given a
finite state interpretation.  This problem is well studied for
the linear case, and compilers exist that will take a rule
formatted somewhat like the SPE style and produce an equivalent
finite state transducer.  Whole constellations of ordered rules or
optimality-theoretic constraints can also be compiled in this way.
However, the compilation of rules and constraints involving
autosegmental structures is still largely un-addressed.

The finite state approaches emphasise the temporal (or left-to-right)
ordering of phonological representations.  In contrast, attribute-value
models emphasise the hierarchical nature of phonological representations.

\subsection{Attribute-value matrices}

The success of attribute-value matrices (AVMs) as a convenient
formal representation for constraint-based approaches to syntax
(see chapter 3),
and concerns about the formal properties of non-linear
phonological information, led some researchers to 
apply AVMs to phonology.  Hierarchical structures can be
represented using AVM nesting, as shown in
(\ref{ex:avm}a), and autosegmental diagrams can be
encoded using AVM indexes, as shown in (\ref{ex:avm}b).

\begin{examples}
\item\label{ex:avm}
\begin{subexamples}
\item
\begin{avm}
\[ onset & \< k \> \\
   rhyme & \[ nucleus & \< u, i \>\\
              coda & \< h \> \] \]
\end{avm}
\item
\begin{avm}
\[ syllable & \< i$_{\@1}$, bu$_{\@2}$, l\textscripta$_{\@3}$, li$_{\@4}$ \> \\
   tone     & \< H$_{\@5}$, L$_{\@6}$, H$_{\@7}$, L$_{\@8}$ \> \\
   associations
            & \{ \<\@1, \@5\>, \<\@2, \@5\>, \<\@3, \@7\>, \<\@4, \@8\> \}
\]
\end{avm}
\end{subexamples}
\end{examples}

AVMs permit re-entrancy by virtue of the numbered indexes, and so
parts of a hierarchical structure can be shared.  For example,
(\ref{ex:sharing}a) illustrates a consonant shared between
two adjacent syllables, for the word \ling{cousin} (this kind
of double affiliation is called \term{ambisyllabicity}).
Example (\ref{ex:sharing}b) illustrates shared structure within
a single syllable \ling{full}, to represent the \term{coarticulation} of
the onset consonant with the vowel.

\begin{examples}
\item\label{ex:sharing}
\begin{subexamples}
\item
\begin{avm}
\[ syllable & \<
  \[ onset & \< k \> \\
     rhyme &
    \[ nucleus & \< \textturnv \>\\
       coda & \< z$_{\@1}$ \> \] \]
  \[ onset & \< \@1 \> \\
     rhyme &
    \[ nucleus & \< \textschwa \>\\
       coda & \< n \> \] \]
\> \]
\end{avm}
\item
\begin{avm}
\[
onset & \[
  consonantal & \[ grave & + \\ compact & -- \] \\
  source & \[ voice & -- \\ continuant & + \] \\
  vocalic & \@1\[ grave & + \\ height & close \] \] \\
rhyme & \[
  nucleus \| vocalic & \@1 \\
  coda & \[
    consonantal & \[ grave & -- \\ compact & -- \] \\
    vocalic & \[ grave & + \\ compact & + \] \\
    source \| nasal & 1 \\
    \]
  \]
\]
\end{avm}
\end{subexamples}
\end{examples}

Given such flexible and extensible representations,
rules and constraints can manipulate and enrich the
phonological information.  Computational implementations
of these AVM models have been used in speech synthesis systems.

\subsection{Computational Tools for Phonological Research}

Once a phonological model is implemented, it ought to be possible
to use the implementation to evaluate theories against data sets.
A phonologist's workbench should help people to `debug' their
analyses and spot errors before going to press with an analysis.
Developing such tools is much more difficult than it might appear.

First, there is no agreed method for modelling non-linear
representations, and each proposal has shortcomings.
Second, processing data sets presents its own set of problems,
having to do with tokenisation, symbols which are ambiguous as
to their featural decomposition, symbols marked as uncertain
or optional, and so on.
Third, some innocuous looking rules and constraints
may be surprisingly difficult to model, and it might only be
possible to approximate the desired behaviour.  Additionally,
certain universal principles and tendencies may be hard to
express in a formal manner.
A final, pervasive problem is that symbolic transcriptions may fail to
adequately reflect linguistically significant
acoustic differences in the speech signal.

Nevertheless, whether the phonologist is
sorting data, or generating helpful tabulations, or
gathering statistics, or searching for a (counter-)example, or
verifying the transcriptions used in a manuscript,
the principal challenge remains a computational one.
Recently, new directed-graph models
(e.g.\ Emu, MATE, Annotation Graphs)
appear to provide good solutions to the first two problems,
while new advances on finite-state models of phonology
are addressing the third problem.
Therefore, we have grounds for confidence that there
will be significant advances on these problems in the near future.

\section*{Further reading and relevant resources}
\label{sec:reading}

The phonology community is served by an excellent journal
{\it Phonology}, published by Cambridge University Press.
Useful textbooks and collections include:
  \citep{Katamba89,FrostKatz92,Kenstowicz94,Goldsmith95,ClarkYallop95,Gussenhoven98,Goldsmith99,Roca99,JurafskyMartin00,HarringtonCassidy00}.
Oxford University Press publishes a series
{\it The Phonology of the World's Languages},
including monographs on
Armenian \citep{Vaux98},
Dutch \citep{Booij95},
English \citep{Hammond99},
German \citep{Wiese96},
Hungarian \citep{Siptar00}.
Kimatuumbi \citep{Odden96},
Norwegian \citep{Kristoffersen96},
Portuguese \citep{Mateus00}, and
Slovak \citep{Rubach93}.
An important forthcoming survey of phonological variation is
the Atlas of North American English \citep{Labov01}.

Phonology is the oldest discipline in linguistics and has a rich
history.  Some historically important works include:
\citep{Joos57,Pike47,Firth48,Bloch48,Hockett55,ChomskyHalle68}.
The most comprehensive history of phonology is \citep{Anderson85}.

Useful resources for phonetics include:
  \citep{Catford88,Laver94,Ladefoged96,Stevens99,IPA99,Ladefoged00,Handke01}, and
the homepage of the International Phonetic Association
\url{http://www.arts.gla.ac.uk/IPA/ipa.html}.
The phonology/phonetics interface is an area of vigorous research,
and the main focus of the {\it Laboratory Phonology} series published by
Cambridge:
  \citep{LabPhon1,LabPhon2,LabPhon3,LabPhon4,LabPhon5}.
Two interesting essays on the relationship between phonetics and phonology
are \citep{Pierrehumbert90,Fleming00}.  Coleman has shown that in
Tashlhiyt Berber (Morocco), where many words appear to have no vowels,
careful phonetic analysis dramatically simplifies the phonological analysis
of syllable structure \citep{Coleman01}.

Important works on the syllable, stress, intonation and tone include
the following:
  \citep{PikePike47,LibermanPrince77,Burzio94,Hayes94,Blevins95,Ladd96,Hirst98,HymanKisseberth98,HulstRitter99}.
Studies of partial specification and redundancy include:
  \citep{Archangeli88,Broe93,Archangeli94}.

Attribute-value and directed graph models for phonological representations
and constraints are described in the following papers and monographs:
  \citep{BirdKlein94,Bird95,Coleman98,Scobbie98,BirdLiberman01,CassidyHarrington01}.

The last decade has seen two major developments in phonology, both falling
outside the scope of this limited chapter.  On the theoretical side, Alan
Prince, Paul Smolensky, John McCarthy and many others have developed a
model of constraint interaction called
{\it Optimality Theory} (OT)
\citep{Archangeli97,Kager99,Tesar00}.
The Rutgers Optimality Archive houses an extensive collection of
OT papers [\url{http://ruccs.rutgers.edu/roa.html}].
On the computational side, the Association for Computational Linguistics
(ACL) has a special interest group in computational phonology (SIGPHON) with
a homepage at \url{http://www.cogsci.ed.ac.uk/sigphon/}.
The organization has held five meetings to date, with proceedings published
by the ACL and many papers available online from the SIGPHON site:
  \citep{SIGPHON-1,SIGPHON-2,SIGPHON-3,SIGPHON-4,SIGPHON-5}.
Another collection of papers was published as a special issue of the
journal {\it Computational Linguistics} in 1994 \citep{Bird94}.
Several PhD theses on computational phonology have appeared:
  \citep{Bird95,Kornai95,Tesar95,CarsonBerndsen97,Walther97,Boersma98,Wareham99,Kiraz00}.
Key contributions to computational OT include the proceedings of the
fourth and fifth SIGPHON meetings, and
\citep{Ellison94b,Tesar95,Eisner97,Karttunen98}.

The sources of data published in this chapter are as follows:
Russian \citep{Kenstowicz79};
Chakosi (Ghana: Language Data Series, ms);
Ulwa \citep[49]{Sproat92}.

\section{Acknowledgements}

I am grateful to D.\ Robert Ladd and Eugene Buckley for comments on an earlier
version of this chapter, and to James Roberts for furnishing me with the
Chakosi data.

\end{document}